\documentclass[10pt,twocolumn,letterpaper]{article}

\usepackage{cvpr}              %

\usepackage{graphicx}
\usepackage{amsmath}
\usepackage{amssymb}
\usepackage{booktabs}
\usepackage{times}
\usepackage{epsfig}
\usepackage{graphicx}
\usepackage{amsmath}
\usepackage{amssymb}
\usepackage{array}
\usepackage{csquotes}
\usepackage[accsupp]{axessibility} %

\usepackage[numbers,sort]{natbib}
\usepackage{xcolor, colortbl}
\usepackage{tabularx}
\usepackage{multirow}
\usepackage{enumitem}
\usepackage{bbm}
\usepackage{pifont}
\usepackage{capt-of}
\usepackage{array,multirow}
\usepackage{booktabs,siunitx,caption}
\usepackage[export]{adjustbox}

\newcolumntype{L}[1]{>{\raggedright\let\newline\\\arraybackslash\hspace{0pt}}m{#1}}
\newcolumntype{C}[1]{>{\centering\let\newline\\\arraybackslash\hspace{0pt}}m{#1}}
\newcolumntype{R}[1]{>{\raggedleft\let\newline\\\arraybackslash\hspace{0pt}}m{#1}}

\def\ie{\emph{i.e.}}
\def\eg{\emph{e.g.}}

\definecolor{Gray}{gray}{0.85}
\definecolor{brown}{rgb}{0.8, 0.2, 0.2}
\definecolor{purple}{rgb}{0.65, 0.1, 0.75}
\definecolor{yellow}{rgb}{0.75, 0.75, 0.0}
\definecolor{orange}{rgb}{1.0, 0.5, 0.2}
\definecolor{green}{rgb}{0, 0.8, 0.2}
\definecolor{red}{rgb}{1.0, 0.0, 0.0}
\definecolor{dblue}{rgb}{0.6, 0.1, 0.9}

\definecolor{grey}{rgb}{0.9, 0.9, 0.9}
\newcommand{\ccol}{\cellcolor{grey}}

\newcommand{\cmark}{\textcolor{red}{\ding{51}}} %
\newcommand{\xmark}{\textcolor{green}{\ding{55}}} %

\newcommand{\bcmark}{\ding{51}}%

\usepackage[pagebackref,breaklinks=true,colorlinks,bookmarks=false]{hyperref}

\usepackage[capitalize]{cleveref}
\crefname{section}{Sec.}{Secs.}
\Crefname{section}{Section}{Sections}
\Crefname{table}{Table}{Tables}
\crefname{table}{Tab.}{Tabs.}

\def\cvprPaperID{6692} %
\def\confName{CVPR}
\def\confYear{2023}

\begin{document}

\title{Improving Cross-Modal Retrieval with Set of Diverse Embeddings}

\author{
Dongwon Kim$^1$ \hspace{12mm} Namyup Kim$^1$ \hspace{12mm} Suha Kwak$^{1,2}$ \hspace{12mm}\\
$^1$Dept. of CSE, POSTECH \hspace{12mm} $^2$ Graduate School of AI, POSTECH\\
{\tt\small \url{https://cvlab.postech.ac.kr/research/DivE/}}
}

\maketitle

\begin{abstract}
Cross-modal retrieval across image and text modalities is a challenging task due to its inherent ambiguity: An image often exhibits various situations, and a caption can be coupled with diverse images.
Set-based embedding has been studied as a solution to this problem. 
It seeks to encode a sample into a set of different embedding vectors that capture different semantics of the sample.
In this paper, we present a novel set-based embedding method, which is distinct from previous work in two aspects. 
First, we present a new similarity function called smooth-Chamfer similarity, which is designed to alleviate the side effects of existing similarity functions for set-based embedding. 
Second, we propose a novel set prediction module to produce a set of embedding vectors that effectively captures diverse semantics of input by the slot attention mechanism.
Our method is evaluated on the COCO and Flickr30K datasets across different visual backbones, where it outperforms existing methods including ones that demand substantially larger computation at inference.

\end{abstract}

\section{Introduction}
\label{sec:intro}

Cross-modal retrieval is the task of searching for data relevant to a query from a database when the query and database have different modalities.
While it has been studied for various pairs of modalities such as video-text~\cite{ging2020coot,bain21frozen,chen2020fine} and audio-text~\cite{chechik2008large,elizalde2019cross}, the most representative setting for the task is the retrieval across image and text modalities~\cite{karpathy2015deep,song2019polysemous,chun2021probabilistic,miech2021thinking}.
A na\"ive solution to cross-modal retrieval
is a straightforward extension of the conventional unimodal retrieval framework~\cite{frome2013devise,faghri2018vse++,jeong2021asmr}, \ie, learning a joint embedding space of the different modalities with known ranking losses (\eg,  contrastive loss~\cite{Chopra2005} and triplet loss~\cite{Schroff2015}).
In this framework, each sample is represented as a single embedding vector and the task reduces to neighbor search on the joint embedding space.

\begin{figure} [!t]
\centering
\includegraphics[width=\columnwidth]{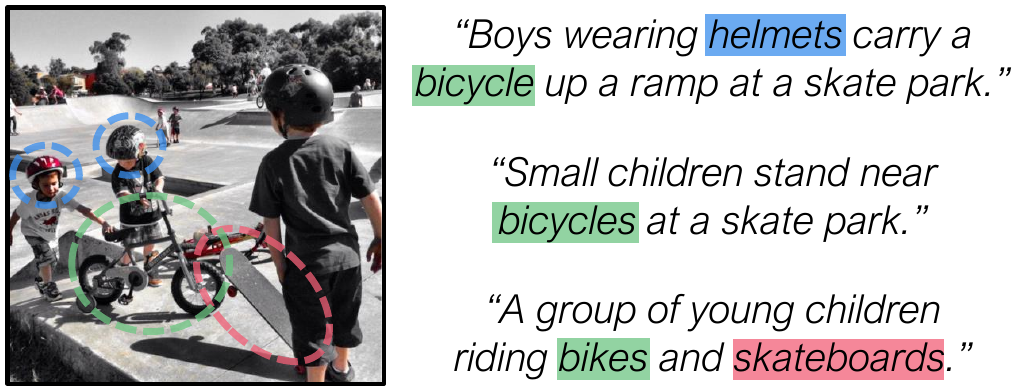}
\caption{
An example of the ambiguity problem introduced in the cross-modal retrieval task;
an image region and a word corresponding to each other are highlighted in the same color.
This example demonstrates that a single image can be coupled with multiple heterogeneous captions.
}
\label{fig:teaser}
\end{figure}

However, this na\"ive approach has trouble in handling the inherent ambiguity of the cross-modal retrieval across image and text modalities~\cite{song2019polysemous,chun2021probabilistic,Thomas2022EmphasizingCS}.
A cause of the ambiguity is the fact that even
a single image often contains various situations and contexts.
Consider an image in Figure~\ref{fig:teaser}, which illustrates a group of children in a skate park. 
One of the captions coupled with it could be about children carrying up a bike, while another may describe the child riding a skateboard. 
Indeed, different local features of the image are matched to different captions. 
Similarly, visual manifestations of a caption could vary significantly as text descriptions are highly abstract.
This ambiguity issue suggests that a sample should be embedded while reflecting its varying semantics in cross-modal retrieval.
Embedding models dedicated to the uni-modal retrieval do not meet this requirement since they represent a sample as a single embedding vector.

Various methods have been studied to mitigate the ambiguity issue of cross-modal retrieval.
Most of them adopt cross-attention networks that directly predict the similarity of an input image-caption pair~\cite{lee2018stacked,miech2021thinking,wei2020universal,desai2021virtex,huang2020pixel,nam2017dual,wang2019camp,zhang2020context,Diao2021SGRAF}.
These models successfully address the ambiguity since they explicitly infer relations across the modalities by drawing attentions on both modalities at once.
However, they inevitably impose a large computation burden on retrieval systems since they demand both image and caption to be processed together for computing their similarity;
all data in a database thus have to be reprocessed whenever a query arrives.
In contrast, methods using separate textual and visual encoders~\cite{karpathy2015deep,faghri2018vse++,gu2018look,huang2018learning} seek to find samples relevant to query through nearest-neighbor search on pre-computed embedding vectors, which are more suitable to nowadays retrieval systems working on huge databases.
However, most of the previous arts in this direction cannot address the ambiguity issue since their encoders return a single embedding vector for a given input.

Recent studies~\cite{song2019polysemous,chun2021probabilistic} have advanced embedding model architectures to tackle the ambiguity issue even with separate encoders for the two modalities.
To be specific, their models compute a set of heterogeneous embedding vectors for a given sample using self-attention layers on top of visual or textual features;
such a set of embedding vectors is called \emph{embedding set} in the remainder of this paper.
Then the ambiguity issue is addressed through elements of the embedding set that encode diverse semantics of input.

Although the set-based embedding models enable a retrieval system to be powerful yet efficient, however, similarity functions used for their training do not consider the ambiguity of the data. 
Hence, training the models with the similarity functions often causes the following two side effects:
(1) \emph{Sparse supervision}--An embedding set most of whose elements remain untrained, or
(2) \emph{Set collapsing}--An embedding set with a small variance where elements do not encode sufficient ambiguity.
Further, self-attention modules used for set prediction in the previous work do not explicitly consider disentanglement between set elements.
These limitations lead to an embedding set whose elements encode redundant semantics of input, which also causes the \emph{set collapsing} and degrades the capability of learned embedding models.

To address the aforementioned limitations of previous work, we propose a novel set-based embedding method for cross-modal retrieval.
The proposed method is distinct from previous work in mainly two aspects.
First, we design a novel similarity function for sets, called smooth-Chamfer similarity, that is employed for both training and evaluation of our model.
In particular, our loss based on the smooth-Chamfer similarity addresses both limitations of the existing similarity functions, \ie, sparse supervision and set collapsing.
Second, we propose a model with a novel set prediction module
motivated by
slot attention~\cite{locatello2020object}.
In the proposed module, learnable embeddings called element slots compete with each other for aggregating input data
while being transformed into an embedding set by progressive update.
Therefore, our model captures the diverse semantic ambiguity of input successfully, with little redundancy between elements of the embedding set.

The proposed method is evaluated and compared with previous work on two realistic cross-modal retrieval benchmarks, 
COCO~\cite{Mscoco} and Flickr30K~\cite{Flickr30k_a}, where it outperforms the previous state of the art in most settings.
In summary, our contribution is three-fold as follows:
\begin{itemize}
   \item We address issues on previous set-based embedding methods by proposing a novel similarity function for sets, named smooth-Chamfer similarity. 
   \item We introduce a slot attention based set prediction module where elements of embedding set iteratively compete with each other for aggregating input data, which can capture semantic ambiguity of input without redundancy.
   \item Our model achieved state-of-the-art performance on the COCO and Flickr30K datasets, two standard benchmarks for cross-modal retrieval.
\end{itemize}

\section{Related Work}
\label{sec:relatedwork}

\noindent \textbf{Cross-modal retrieval:}
Previous work can be categorized into two classes, methods using two separate encoders and those using a cross-attention network.
Models of the first class consist of disjoint visual and textual encoders and learn an embedding space where embedding vectors of matching pairs are located nearby.
Retrieval is then performed by finding nearest neighbors of query on the embedding space, which is efficient since their embedding vectors are pre-computed.
To improve the quality of the embedding space,
loss functions~\cite{faghri2018vse++,chun2021probabilistic}, model architectures~\cite{huang2018learning,li2019visual,Chen2021gpo}, and embedding methods~\cite{song2019polysemous,chun2021probabilistic} have been proposed. 
Though the separate encoders enable simple and fast retrieval, 
they often failed to handle the inherent ambiguity of cross-modal retrieval, leading to inferior retrieval performance. 
On the other hand, models of the other class adopt a cross-attention network that directly predicts the similarity score between an image-caption pair~\cite{lee2018stacked,wei2020universal,wang2019camp,zhang2020context,Diao2021SGRAF,Zhang_2022_CVPR}. 
Though these methods make their models explicitly consider relations across modalities, they are not suitable for real-world retrieval scenarios due to the large computation burden imposed during evaluation. 
We thus focus on improving the separate encoders in this paper.

\begin{figure*} [!t]
\centering
\includegraphics[width=\textwidth]{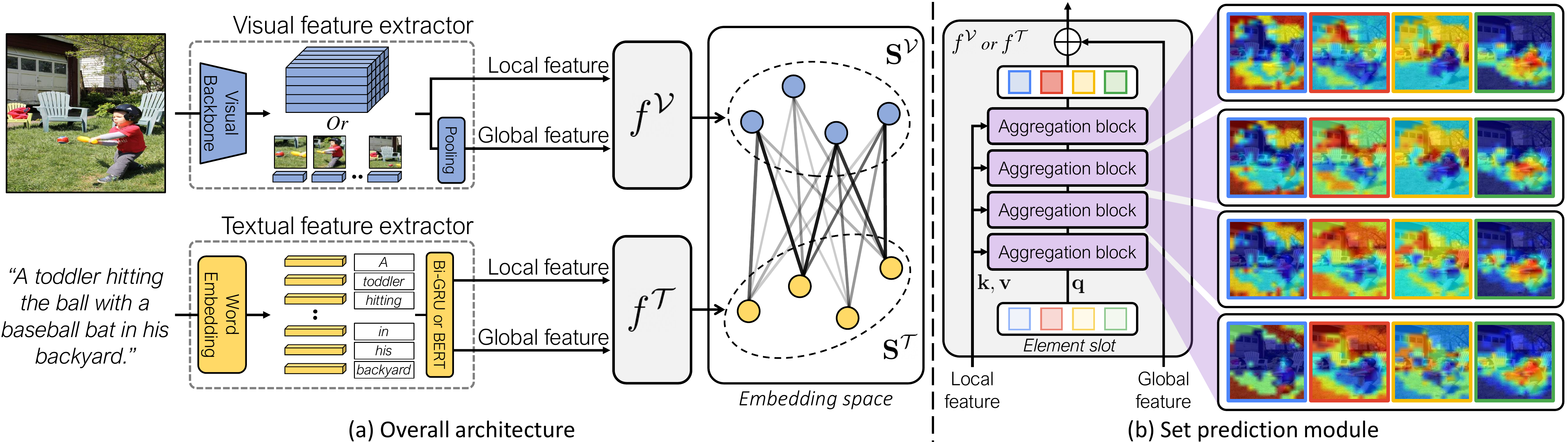}
\caption{
An overview of our model. 
(a) The overall framework of our model.
The model consists of three parts: visual feature extractor, textual feature extractor, and set-prediction modules $f^\mathcal{V}$ and $f^\mathcal{T}$. 
First, the feature extractors of each modality extract local and global features from input samples. Then, the features are fed to the set prediction modules to produce embedding sets $\mathbf{S}^\mathcal{V}$ and $\mathbf{S}^\mathcal{T}$. 
The model is trained with the loss using our smooth-Chamfer similarity.
(b) Details of our set prediction module and attention maps that slots of each iteration produce.
A set prediction module consists of multiple aggregation blocks that share their weights.
Note that $f^\mathcal{V}$ and $f^\mathcal{T}$ have the same model architecture.
} 
\label{fig:overview}
\end{figure*}

\noindent \textbf{Embedding beyond vector representation:}
Most of existing work projects each sample into a single embedding vector~\cite{faghri2018vse++,gu2018look,huang2018learning}.
However, it has been empirically proven that such single vector representation is not sufficient to deal with inherent ambiguity of data~\cite{vilnis2014word,shi2019probabilistic}.
To mitigate this issue for the cross-modal retrieval task,
PVSE~\cite{song2019polysemous} and PCME~\cite{chun2021probabilistic} introduce novel embedding methods.
To be specific, PVSE represents each sample as a set of embedding vectors and PCME introduces probabilistic embedding where each sample is represented as a set of vectors sampled from a normal distribution.

\noindent \textbf{Slot attention:}
Recently, slot attention~\cite{locatello2020object} is proposed to learn object-centric representation, which is especially beneficial for tasks that require perception of objectness, such as object discovery and set property prediction.
Slots, which are embedding vectors sampled from a random distribution, compete with each other for explaining the input in an iterative manner.
This attention mechanism enables the final outputs to encode heterogeneous semantics that appear in the input without any explicit supervision.
To produce informative embedding sets with sufficient within-set variance, our set prediction module employs the slot attention mechanism.
However, the ability of slot attention to discover individual objects is only verified on synthetic datasets~\cite{johnson2017clevr} and is known to fail on real-world images~\cite{chang2022object}.
We make three architectural modifications to resolve this issue: (1) using learnable embeddings for initial element slots instead of random vectors, (2) replacing GRU~\cite{cho2014learning} with a residual sum, and (3) adding a global feature to the final element slots. 
We will detail these differences in Section~\ref{subsec:method_set_pred}.
Without them, we observed that the loss does not converge, and thus training fails. 

\section{Proposed Method}
\label{sec:method}

This section first describes the overall model architecture and elaborates on the proposed set prediction module.
Then we introduce our smooth-Chamfer similarity and illustrate how it differs from existing set similarity functions.
Finally, details of the training and inference of our model are provided.

\subsection{Overall Model Architecture}
\label{subsec:method_arch}
The overall framework of our method is illustrated in Figure~\ref{fig:overview}. Our model architecture consists of a visual feature extractor, a textual feature extractor, and a set prediction module of each modality: $f^\mathcal{V}$ and $f^\mathcal{T}$. 
The feature extractors have two branches that compute local features and global features from the input sample, respectively. 
The extracted features are given as input to the set prediction modules, each of which fuses local and global features to encode an embedding set.
For the visual and textual feature extractors, we followed the conventional setting of the previous work~\cite{song2019polysemous,chun2021probabilistic,Chen2021gpo}.

\vspace{1mm} \noindent \textbf{Visual feature extractor:}
We consider two different types of visual feature extractors for fair comparisons with previous work. 
One employs a flattened convolutional feature map as local features and their average pooled feature as the global feature.
The other uses ROI features of a pretrained object detector as local features and their max pooled feature as the global feature.
In any of these cases, the local and global features are denoted by $\psi^{\mathcal{V}}(\mathbf{x}) \in \mathbb{R}^{N \times D}$ and $\phi^{\mathcal{V}}(\mathbf{x}) \in \mathbb{R}^{D}$, respectively, where $\mathbf{x}$ is input image.

\vspace{1mm} \noindent \textbf{Textual feature extractor:}
For the textual feature, we also employ two different types of extractors: bi-GRU~\cite{cho2014learning} and BERT~\cite{devlin2018bert}.
When using bi-GRU, for $L$-words input caption $\mathbf{y}$, we take GloVe~\cite{pennington2014glove} word embedding of each word as a local feature $\psi^{\mathcal{T}}(\mathbf{y}) \in \mathbb{R}^{L \times 300}$. 
Then we apply bi-GRU with $D$ dimensional hidden states on the top of the $\psi^{\mathcal{T}}(\mathbf{y})$. The last hidden state is used as the global feature $\phi^{\mathcal{T}}(\mathbf{y}) \in \mathbb{R}^{D}$.
Similarly, when using BERT, output hidden states of BERT and their max pooled feature are used as a $\psi^{\mathcal{T}}(\mathbf{y}) \in \mathbb{R}^{L \times D}$ and $\phi^{\mathcal{T}}(\mathbf{y}) \in \mathbb{R}^{D}$, respectively.

\subsection{Set Prediction Module}
\label{subsec:method_set_pred}

Elements of an embedding set should encode heterogeneous semantics that appear in the input data. Otherwise, it will degenerate into a set with small variations that cannot handle the ambiguity of the input.

Inspired by slot attention~\cite{locatello2020object}, we introduce an aggregation block, where \textit{element slots} compete with each other for aggregating input data.
The slots are progressively updated to capture various semantics of the input through multiple iterations of the aggregation block, and then the slots are fused with a global feature to be used as elements of the output embedding set.
In this manner, the proposed module produces an embedding set whose elements encode substantially different semantics, meanwhile preserving the global context of the input data.
In this section, we only present how the visual set prediction module $f^\mathcal{V}$ works; $f^\mathcal{T}$ has the same model architecture and works in the same manner with $f^\mathcal{V}$.

As shown in Figure~\ref{fig:overview}(b), the proposed set prediction module consists of multiple aggregation blocks, which share their weights.
We define the initial element slots as learnable embedding vectors $\mathbf{E}^{0} \in \mathbb{R}^{K \times D}$, where $K$ is the cardinality of the embedding set.
Then, the element slots of the $t$-th iteration $\mathbf{E}^{t}$ are computed by
\begin{align}
    \mathbf{E}^t = \mathrm{AggBlock}(\psi^\mathcal{V}(\mathbf{x}); \mathbf{E}^{t-1}) \in \mathbb{R}^{K \times D}.
\end{align}
The aggregation block first layer-normalizes~\cite{ba2016layer} inputs and then linearly projects input local feature $\psi^\mathcal{V}(\mathbf{x})$ to $\mathbf{k} \in \mathbb{R}^{N \times D_h}$ and $\mathbf{v} \in \mathbb{R}^{N \times D_h}$, and $\mathbf{E}^{t-1}$ to $\mathbf{q} \in \mathbb{R}^{K \times D_h}$.
\footnote{When using convolutional features for visual feature extraction, we add sinusoidal positional encoding before projecting $\psi^\mathcal{V}(\mathbf{x})$ to $\mathbf{k}$.}
Then, the attention map $A$ between $\psi^\mathcal{V}(\mathbf{x})$ and $\mathbf{E}^{t-1}$ is obtained via
\begin{align}
    A_{n,k} = \frac{e^{M_{n,k}}}{\sum_{i=1}^K e^{M_{n,i}}}, \text{ where  } M = \frac{\mathbf{k}\mathbf{q}^{\top}}{\sqrt{D_h}}.
\end{align}
It is worth noting that the attention map is normalized over \textit{slots}, not \textit{keys} as in the transformer~\cite{vaswani2017attention}.
Since this way of normalization lets slots compete with each other, each slot attends to nearly disjoint sets of local features, and these sets will correspond to the distinctive semantics of the input.

Using the obtained attention map, we update the element slot with the weighted mean of local features and then feed it to a multi-layer perceptron (MLP) with layer normalization, residual connection, and GELU~\cite{hendrycks2016gaussian} activation:
\begin{align}
    \hat{A}_{n,k} = \frac{A_{n,k}}{\sum_{i=1}^N A_{i,k}}, \text{    } \overline{\mathbf{E}^{t}} = \hat{A}^{\top} \mathbf{v} W_o + \mathbf{E}^{t-1}, 
\end{align}
\begin{align}
    \mathbf{E}^{t} = \mathrm{AggBlock}(\psi^\mathcal{V}(\mathbf{x}); \mathbf{E}^{t-1}) = \mathrm{MLP}(\overline{\mathbf{E}^{t}}) + \overline{\mathbf{E}^{t}},
\end{align}
where $W_0 \in \mathbb{R}^{D_h \times D}$ is a learnable linear projection.
The output of the aggregation block is used as an element slot for the $t$-th iteration. As shown in Figure~\ref{fig:overview}, attention maps that slots of early stage produce are sparse and noisy, but refined to aggregate local features with distinctive semantics as proceeds.

Finally, at $T$-th iteration, the model predicts an embedding set $\mathbf{S}$ by adding the global feature $\phi^{\mathcal{V}}(\mathbf{x})$ to each element slot $\mathbf{E}^{T}$:
\begin{align}
    \mathbf{S} = \mathrm{LN}(\mathbf{E}^{T}) + \big[\mathrm{LN}(\phi^{\mathcal{V}}(\mathbf{x}))\stackrel{\scriptscriptstyle \times K}{\cdots}\big],
\end{align}
where $\mathrm{LN}$ is a layer normalization, and $\big[\phi^{\mathcal{V}}(\mathbf{x})\stackrel{\scriptscriptstyle \times K}{\cdots}\big] \in \mathbb{R}^{K \times D}$ is $K$ repetitions of $\phi^{\mathcal{V}}(\mathbf{x})$.
The module benefits from the global feature when treating samples with little ambiguity since the global feature allows them to be represented as an embedding set of small within-set variance.

\subsection{Smooth-Chamfer Similarity}
\label{subsec:method_similarity}

\begin{figure} [!t]
\centering
\includegraphics[width=0.98\linewidth]{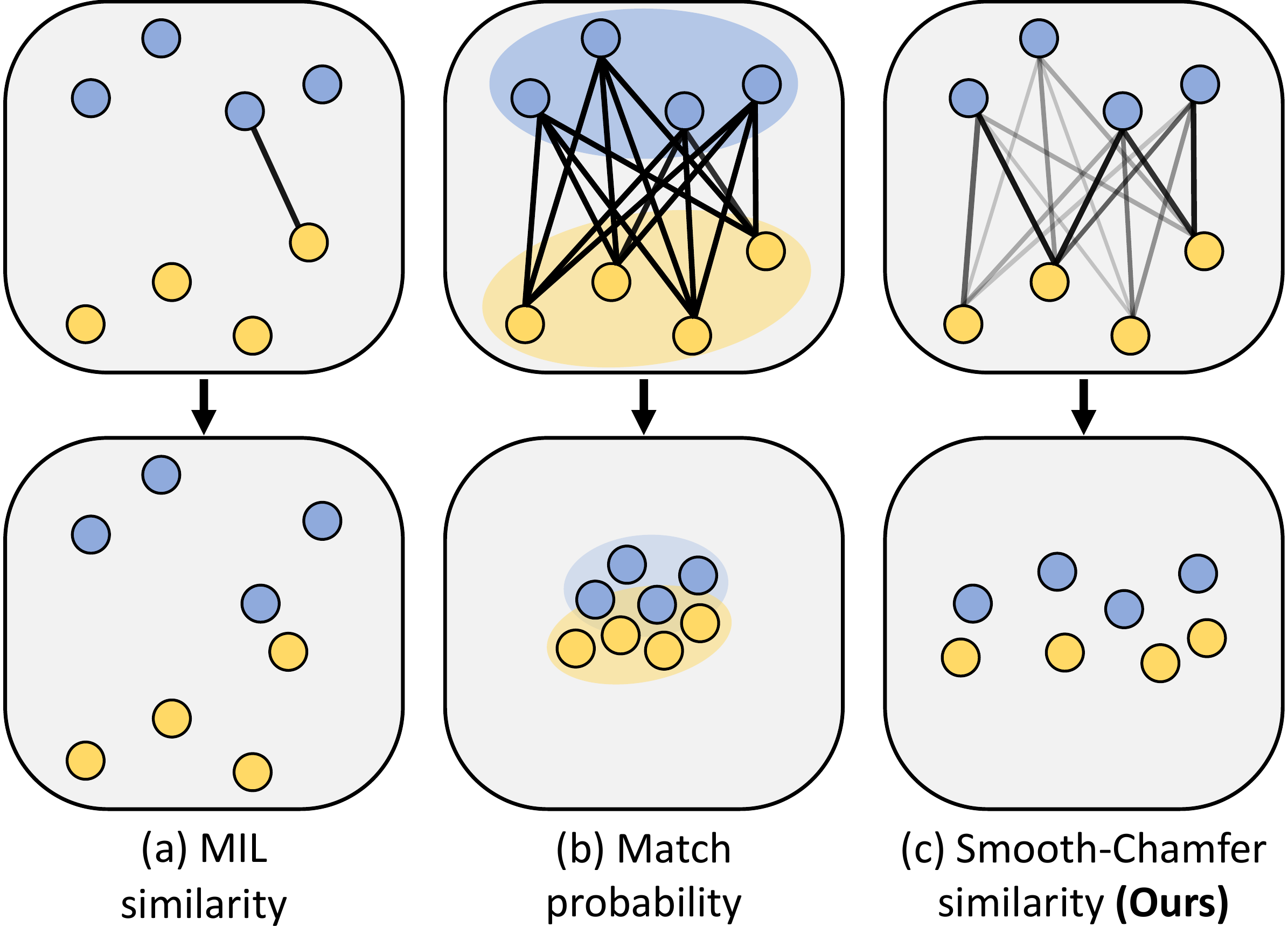}
\caption{
Comparison between our similarity function and existing ones used for embedding set.
Illustrations of embedding space before and after optimizing each similarity function are presented. 
Presented two sets are matching pairs, and thus optimized to maximize their similarity.
Lines indicate the association that similarity functions consider, where their intensities represent the weights given to each association.
} 
\label{fig:similarities}
\end{figure}

Our smooth-Chamfer similarity is proposed to resolve the drawbacks of the existing set-based embedding approaches.
Before presenting smooth-Chamfer similarity, we first review the similarity functions used in PVSE~\cite{song2019polysemous} and PCME~\cite{chun2021probabilistic}. 
Figure~\ref{fig:similarities} illustrates embedding spaces trained with these set similarity functions.
Let $\mathbf{S}_1$ and $\mathbf{S}_2$ be the embedding sets, which are the set of vectors. $c(x,y)$ denotes the cosine similarity between vectors $x$ and $y$.

PVSE adopts the multiple instance learning (MIL) framework~\cite{MIL_Dietterich} during training and inference. Its similarity function, which we call MIL similarity, is given by $s_{\mathrm{MIL}}(\mathbf{S}_1, \mathbf{S}_2) = \max_{x \in \mathbf{S}_1, y \in \mathbf{S}_2} c(x,y)$.
MIL similarity only takes account of the closest pair of elements, as shown in Figure~\ref{fig:similarities}(a).
While this behavior simplifies similarity measurement, MIL similarity suffers from the \emph{sparse supervision} problem. 
In other words, the majority of elements are not sampled as the closest pair and thus remain untrained.

On the other hand, PCME uses the match probability (MP) as a similarity function, which is defined as $s_{\mathrm{MP}}(\mathbf{S}_1, \mathbf{S}_2) = \sum_{x \in \mathbf{S}_1, y \in \mathbf{S}_2} \sigma(\alpha c(x,y) + \beta),$ where $\alpha$ and $\beta$ are learnable parameters and $\sigma$ is the sigmoid function.
MP takes average on every pairwise distance between elements.
Though MP resolves the sparse supervision, it introduces the \emph{set collapsing} problem.
As presented in Figure~\ref{fig:similarities}(b), directly optimizing MP leads to a collapsed embedding set since elements of embedding sets pull each other without the consideration of their relative proximity.
To mitigate this issue, MP requires a specialized loss function~\cite{chun2021probabilistic,oh2018modeling}, which makes MP incompatible with other losses or frameworks.

The aforementioned problems stem from the way how similarity functions associate set elements.
Associating only the nearest pair of elements like MIL leads to insufficient supervisory signals, while equally taking account of every possible association like MP results in a collapsed set.
Unlike these similarity functions, our smooth-Chamfer similarity associates every possible pair with different degrees of weight, as illustrated in Figure~\ref{fig:similarities}(c). 
Smooth-Chamfer similarity is formulated as
\begin{equation}
\begin{split}
s_{\mathrm{SC}}(\mathbf{S}_1, \mathbf{S}_2) = \frac{1}{2\alpha|\mathbf{S}_1|} \sum_{x \in \mathbf{S}_1} \log\bigg(\sum_{y \in \mathbf{S}_2}e^{\alpha c(x,y)}\bigg) \\
+ \frac{1}{2\alpha|\mathbf{S}_2|} \sum_{y \in \mathbf{S}_2} \log\bigg(\sum_{x \in \mathbf{S}_1}e^{\alpha c(x,y)}\bigg),
\end{split}
\label{eq:smooth_chamfer_sim}
\end{equation}
where $|\cdot|$ is the set cardinality and $\alpha > 0$ is a scaling parameter.
Using Log-Sum-Exp, it can be reformulated in a more interpretable form as shown below:
\begin{equation}
\small
\begin{split}
s_{\mathrm{SC}}(\mathbf{S}_1, \mathbf{S}_2) = \frac{1}{2\alpha|\mathbf{S}_1|} \sum_{x \in \mathbf{S}_1} \underset{y \in \mathbf{S}_2}{\mathrm{LSE}} \bigg(\alpha c(x,y)\bigg)\\
+ \frac{1}{2\alpha|\mathbf{S}_2|} \sum_{y \in \mathbf{S}_2} \underset{x \in \mathbf{S}_1}{\mathrm{LSE}} \bigg(\alpha c(x,y)\bigg).
\end{split}
\label{eq:smooth_chamfer_sim_lse}
\end{equation}
LSE indicates Log-Sum-Exp, which is the smooth approximation of the max function.
Thanks to the property of LSE, smooth-Chamfer similarity softly assigns elements in one set to those in the other set, where the weight for a pair of elements is determined by their relative proximity.
Hence, smooth-Chamfer similarity can provide dense supervision like MP but without the collapsing issue.

By replacing LSE with the max function, we can consider 
a \textit{non-smooth} version
of smooth-Chamfer similarity, which we refer to as Chamfer similarity.
Chamfer similarity clears up the problem of set collapsing by assigning individual elements of $\mathbf{S}_1$ to their closest neighbor in $\mathbf{S}_2$, and vice-versa. 
However, unlike smooth-Chamfer similarity, it does not provide supervision that is dense as MP since most associations are not considered.

The behavior of smooth-Chamfer similarity can be clearly demonstrated by its gradient with respect to $c(x,y)$, which is given by
\begin{equation}
\begin{split}
\frac{\partial s_{\mathrm{SC}}(\mathbf{S}_1, \mathbf{S}_2)}{\partial c(x',y')} = \frac{1}{2|\mathbf{S}_1|} \frac{e^{\alpha c(x',y')}}{\sum_{y \in \mathbf{S}_2} e^{\alpha c(x',y)}} \\
+ \frac{1}{2|\mathbf{S}_2|} \frac{e^{\alpha c(x',y')}}{\sum_{x \in \mathbf{S}_1} e^{\alpha c(x,y')}},
\end{split}
\label{eq:smooth_chamfer_sim_grad}
\end{equation}
where $x'$ and $y'$ are elements of $\mathbf{S}_1$ and $\mathbf{S}_2$, respectively.
One can see that the gradient is the sum of two relative similarity scores, suggesting that the gradient for $c(x',y')$ is emphasized when $x'$ and $y'$ are close to each other.
Using the weighting scheme based on the relative proximity,
we can give 
denser supervision during training, while preserving sufficient within-set variance.

\subsection{Training and Inference}
\label{subsec:training_inference}
\noindent \textbf{Training:} 
Our model is trained with the objective functions presented in~\cite{song2019polysemous}, which consists of the triplet loss, diversity regularizer, and Maximum Mean Discrepancy (MMD)~\cite{gretton2006kernel} regularizer. 
Following previous work~\cite{faghri2018vse++,song2019polysemous,wei2020universal}, we adopt the triplet loss with hard negative mining. 
Let $B=\{(S^\mathcal{V}_i, S^\mathcal{T}_i)\}^{N}_{i=1}$ be a batch of embedding sets. %
Then the triplet loss is given by
\begin{align}
\begin{split}
    \mathcal{L}_{\mathrm{tri}}(B) = \sum_{i=1}^{N} \max_{j} [\delta + s_{\mathrm{SC}}(\mathbf{S}^{\mathcal{V}}_i, \mathbf{S}^{\mathcal{T}}_j) - s_{\mathrm{SC}}(\mathbf{S}^{\mathcal{V}}_i, \mathbf{S}^{\mathcal{T}}_i)]_+ \\
    + \sum_{i=1}^N \max_{j} [\delta + s_{\mathrm{SC}}(\mathbf{S}^{\mathcal{T}}_i, \mathbf{S}^{\mathcal{V}}_j) - s_{\mathrm{SC}}(\mathbf{S}^{\mathcal{T}}_i, \mathbf{S}^{\mathcal{V}}_i)]_+,
\end{split}
\end{align}
where $[\cdot]_+$ indicates the hinge function and $\delta > 0$ is a margin.
The MMD regularizer minimizes MMD between embedding sets of image and text. It prevents embeddings of different modalities from diverging at the early stage of training.
The diversity regularizer penalizes similar element slots, which helps the model produce a set of diverse embeddings. The two regularizers are formulated as $\mathcal{L}_{\mathrm{mmd}}=\mathrm{MMD}(B^\mathcal{V}, B^\mathcal{T})$ and $\mathcal{L}_{\mathrm{div}} = \sum_{x, x' \in \mathbf{E}^{T}} e^{-2||x-x'||^2_2}$, where $B^\mathcal{V}$ and $B^\mathcal{T}$ denote subsets of batch $B$ consisting of each modality embeddings.

\vspace{1mm} \noindent \textbf{Inference:}
We pre-compute the embedding set of size $K$ for every sample in the database.
Then, the sample most relevant to a query is retrieved
via nearest-neighbor search on embedding sets, using smooth-Chamfer similarity.

\section{Experiments}
\label{sec:experiment}

\begin{table*}[t]
\centering
\scalebox{0.955}{
\footnotesize
\begin{tabular}{l|c|ccc|ccc|c|ccc|ccc|c}
\toprule
    \multicolumn{1}{l}{\multirow{3}{*}[-1.6em]{Method}} 
    & \multicolumn{1}{c|}{\multirow{3}{*}[-1.6em]{CA}} 
    & \multicolumn{7}{c|}{1K Test Images} 
    & \multicolumn{7}{c}{5K Test Images} \\ \midrule
    \multicolumn{1}{c|}{}
    & 
    & \multicolumn{3}{c|}{Image-to-Text} 
    & \multicolumn{3}{c|}{Text-to-Image}
    & \multicolumn{1}{c|}{\multirow{2}{*}{\textsc{RSUM}}}
    & \multicolumn{3}{c|}{Image-to-Text} 
    & \multicolumn{3}{c|}{Text-to-Image}
    & \multicolumn{1}{c}{\multirow{2}{*}{RSUM}} \\
    \multicolumn{1}{c|}{}
    &
    & R@1 & R@5 & R@10 & R@1 & R@5 & R@10 &
    & R@1 & R@5 & R@10 & R@1 & R@5 & R@10 & \\
\midrule %
\multicolumn{16}{l}{\textit{\textbf{ResNet-152 + Bi-GRU}}} \\ \midrule
VSE++~\cite{faghri2018vse++} & \xmark
    & 64.6 & 90.0 & 95.7& 52.0 & 84.3 & 92.0 & 478.6
    & 41.3 & 71.1 & 81.2 & 30.3 & 59.4 & 72.4 & 355.7\\
PVSE~\cite{song2019polysemous} & \xmark
    & 69.2 & 91.6 & 96.6 & 55.2 & 86.5 & 93.7 & 492.8
    & 45.2 & 74.3 & 84.5 & 32.4 & 63.0 & 75.0 & 374.4\\
PCME~\cite{chun2021probabilistic} & \xmark
    & 68.8 & - & - & 54.6 & - & - & - & 44.2 & - & - & 31.9 & - & - & - \\
\ccol \textbf{Ours} & \ccol \xmark
   & \ccol 70.3 & \ccol 91.5 & \ccol 96.3 & \ccol 56.0 & \ccol 85.8 & \ccol 93.3 & \ccol \textbf{493.2}
   & \ccol 47.2 & \ccol 74.8 & \ccol 84.1 & \ccol 33.8 & \ccol 63.1 & \ccol 74.7 & \ccol \textbf{377.7} \\

\midrule %
\multicolumn{16}{l}{\textit{\textbf{Faster R-CNN + Bi-GRU}}} \\ \midrule

SCAN$^\dagger$~\cite{lee2018stacked} & \cmark
    & 72.7& 94.8& 98.4& 58.8& 88.4& 94.8& 507.9
    & 50.4& 82.2& 90.0& 38.6& 69.3& 80.4& 410.9\\
    
VSRN$^\dagger$~\cite{li2019visual} & \xmark
    & 76.2 & 94.8 & 98.2 & 62.8 & 89.7 & 95.1 & 516.8
    & 53.0 & 81.1 & 89.4 & 40.5 & 70.6 & 81.1 & 415.7 \\

CAAN~\cite{zhang2020context} & \cmark
    & 75.5 & 95.4 & 98.5 & 61.3 & 89.7 & 95.2 & 515.6
    & 52.5 & 83.3 & 90.9 & 41.2 & 70.3 & 82.9 & 421.1 \\

IMRAM$^\dagger$~\cite{chen2020imram} & \cmark
    & 76.7 & 95.6 & 98.5 & 61.7 & 89.1 & 95.0 & 516.6 
    & 53.7 & 83.2 & 91.0 & 39.7 & 69.1 & 79.8 & 416.5\\
    
SGRAF$^\dagger$~\cite{Diao2021SGRAF} & \cmark
    & 79.6 & 96.2 & 98.5 & 63.2 & 90.7 & 96.1 & 524.3 
    & 57.8 & - & 91.6 & 41.9 & - & 81.3 & - \\

VSE$_\infty$~\cite{Chen2021gpo} & \xmark
    & 78.5 & 96.0 & 98.7 & 61.7 & 90.3 & 95.6 & 520.8 
    & 56.6 & 83.6 & 91.4 & 39.3 & 69.9 & 81.1 & 421.9\\

NAAF$^\dagger$~\cite{Zhang_2022_CVPR} & \cmark
    & 80.5 & 96.5 & 98.8 & 64.1 & 90.7 & 96.5 & 527.2 
    & 58.9 & 85.2 & 92.0 & 42.5 & 70.9 & 81.4 & 430.9 \\

\ccol \textbf{Ours} & \ccol  \xmark
   & \ccol 79.8  & \ccol  96.2 & \ccol 98.6 & \ccol 63.6 & \ccol 90.7 & \ccol 95.7 & \ccol 524.6
   & \ccol 58.8 & \ccol  84.9 & \ccol 91.5 & \ccol 41.1 & \ccol 72.0 & \ccol 82.4 & \ccol 430.7 \\
   
\ccol \textbf{Ours$^\dagger$} & \ccol  \xmark
   & \ccol 80.6 & \ccol 96.3  & \ccol 98.8 & \ccol 64.7 & \ccol 91.4 & \ccol 96.2 & \ccol \textbf{528.0}
   & \ccol 60.4 & \ccol 86.2  & \ccol 92.4 & \ccol 42.6 & \ccol 73.1 & \ccol 83.1 & \ccol \textbf{437.8} \\

\midrule %
\multicolumn{16}{l}{\textit{\textbf{ResNeXt-101 + BERT}}} \\ \midrule
VSE$_\infty$~\cite{Chen2021gpo} & \xmark
    & 84.5 & 98.1 & 99.4 & 72.0 & 93.9 & 97.5 & 545.4 
    & 66.4 & 89.3 & 94.6 & 51.6 & 79.3 & 87.6 & 468.9  \\
VSE$_\infty$$^\dagger$~\cite{Chen2021gpo} & \xmark
    & 85.6 & 98.0 & 99.4 & 73.1 & 94.3 & 97.7 & 548.1 
    & 68.1 & 90.2 & 95.2 & 52.7 & 80.2 & 88.3 & 474.8 \\
\ccol \textbf{Ours} & \ccol \xmark
   & \ccol  86.3 & \ccol 97.8  & \ccol 99.4 & \ccol 72.4 & \ccol 94.0 & \ccol 97.6 & \ccol 547.5
   & \ccol 69.1 & \ccol  90.7 & \ccol 95.6 & \ccol 52.1 & \ccol 79.6 & \ccol 87.8 & \ccol 474.9 \\
\ccol \textbf{Ours$^\dagger$} & \ccol \xmark
   & \ccol  86.6& \ccol  98.2 & \ccol 99.4 & \ccol 73.4 & \ccol 94.5 & \ccol 97.8 & \ccol \textbf{549.9}
   & \ccol  71.0& \ccol   91.8& \ccol  96.3& \ccol  53.4& \ccol 80.9 & \ccol 88.6 & \ccol \textbf{482.0} \\
\bottomrule %
\end{tabular}
}
\caption{
Recall@K (\%) and RSUM on the COCO dataset. 
Evaluation results on both 1K test setting (average of 5-fold test dataset) and 5K test setting are presented.
The best RSUM scores are marked in bold. CA and $\dagger$ indicate models using cross-attention and ensemble models of two hypotheses, respectively.
}
\vspace{-5mm}
\label{tab:coco_comparison} 
\end{table*}

\begin{table}[t]
\centering
\scalebox{0.71}{
    \begin{tabular}{l|c|ccc|ccc|c}
    \toprule

    \multirow{2}{*}{Method}
    & \multirow{2}{*}{CA}
    & \multicolumn{3}{c|}{Image-to-text} 
    & \multicolumn{3}{c|}{Text-to-image} 
    & \multirow{2}{*}{RSUM} \\
    
    & & R@1 & R@5 & R@10 & R@1 & R@5 & R@10 & \\ \midrule
    
    \multicolumn{9}{l}{\textit{\textbf{ResNet-152 + Bi-GRU}}} \\ \midrule
    
    VSE++ & \xmark& 52.9& 80.5& 87.2& 39.6& 70.1& 79.5& 409.8\\
    PVSE$^{*}$ & \xmark& 59.1& 84.5& 91.0& 43.4& 73.1& 81.5& 432.6\\
    PCME$^{*}$ & \xmark& 58.5& 81.4& 89.3& 44.3& 72.7 & 81.9& 428.1\\
    \ccol \textbf{Ours} & \ccol\xmark & \ccol 61.8& \ccol 85.5& \ccol 91.1& 
                        \ccol 46.1& \ccol 74.8& \ccol 83.3& \ccol \textbf{442.6}\\

    \midrule %
    \multicolumn{9}{l}{\textit{\textbf{Faster R-CNN + Bi-GRU}}} \\ \midrule
        
    SCAN$^\dagger$  & \cmark & 67.4& 90.3& 95.8& 48.6& 77.7& 85.2& 465.0\\
    
    VSRN$^\dagger$  & \xmark & 71.3& 90.6& 96.0& 54.7& 81.8& 88.2& 482.6\\

    CAAN& \cmark & 70.1 & 91.6 & 97.2 & 52.8 & 79.0 & 87.9 &  478.6 \\
    
    IMRAM$^\dagger$& \cmark & 74.1& 93.0& 96.6& 53.9& 79.4& 87.2& 484.2\\
    
    SGRAF$^\dagger$ & \cmark &
        77.8& 94.1& 97.4& 58.5& 83.0& 88.8& 499.6\\

    VSE${_\infty}$ & \xmark & 
        76.5& 94.2& 97.7& 56.4& 83.4& 89.9& 498.1\\
        
    NAAF$^\dagger$ & \cmark &
        81.9& 96.1& 98.3& 61.0& 85.3& 90.6& \textbf{513.2}\\
        
    \ccol \textbf{Ours} & \ccol\xmark & \ccol 77.8& \ccol 94.0& \ccol 97.5& 
                            \ccol 57.5& \ccol 84.0& \ccol 90.0& \ccol 500.8\\
                            
    \ccol \textbf{Ours$^\dagger$} & \ccol\xmark & \ccol 80.9& \ccol 94.7& \ccol 97.6& 
                            \ccol 59.4& \ccol 85.6& \ccol 91.1& \ccol 509.3\\

    \midrule %
    \multicolumn{9}{l}{\textit{\textbf{ResNeXt-101 + BERT}}} \\ \midrule
    VSE${_\infty}$ & \xmark& 88.4& 98.3& 99.5& 74.2& 93.7& 96.8& 550.9 \\
    VSE${_\infty}$$^\dagger$  &\xmark & 88.7& 98.9& 99.8& 76.1& 94.5& 97.1& 555.1\\
    \ccol \textbf{Ours} & \ccol\xmark& \ccol 88.8& \ccol 98.5& \ccol 99.6& 
                            \ccol 74.3& \ccol 94.0& \ccol 96.7& \ccol 551.9\\
    \ccol \textbf{Ours$^\dagger$} & \ccol\xmark&\ccol 90.6& \ccol 99.0& \ccol 99.6&
                            \ccol 75.9& \ccol 94.7& \ccol 97.3& \ccol \textbf{557.1}\\
                            
    \bottomrule
    \end{tabular}
}
\caption{
Recall@K(\%) and RSUM on the Flickr30K dataset. 
CA, $\dagger$, and * indicate models using cross-attention, ensemble models of two hypotheses, and models we reproduce, respectively.
}
\vspace{-5mm}
\label{tab:f30k_comparison}
\end{table}

\subsection{Datasets and Evaluation Metric}
We validate the effectiveness of our method on COCO~\cite{Mscoco} and Flickr30K~\cite{Flickr30k_a} datasets. In both datasets, we follow the split proposed by~\cite{karpathy2015deep}.
COCO dataset consists of a train split of 113,287 images, a validation split of 5,000 images, and a test split of 5,000 images. 
Retrieval results of our method on 1K test setting and 5K test setting are both reported, following~\cite{karpathy2015deep}. 
In the 1K test setting, the average retrieval performance on the 5-fold test split is reported.
For the Flickr30K dataset, we use 28,000 images for training, 1,000 images for validation, and 1,000 images for testing.
We report the retrieval results on the Flick30K using a test split of 1,000 images.
In both datasets, each image is given with five matching captions.
For evaluation, we use the Recall@$K$, which is the percentage of the queries that have matching samples among top-$K$ retrieval results. 
Following~\cite{Chen2021gpo}, We also report the RSUM, which is the sum of the Recall@$K$ at $K \in \{1, 5, 10\}$ in the image-to-text and text-to-image retrieval settings.

\vspace{3mm}
\subsection{Implementation Details}
\label{subsec:exp_impl}
\noindent \textbf{Feature extractor:}
For the visual feature extractor, convolutional visual features are obtained by applying $1 \times 1$ convolution to the last feature map of CNN.
We obtain ROI visual features by feeding the pre-extracted features~\cite{anderson2018bottom} from a Faster R-CNN~\cite{faster_rcnn} to the 2-layer MLP with residual connection, following~\cite{Chen2021gpo}. 
In every model, we set $D$ to 1024 and $K$ to 4. 

\begin{figure*} [!t]
\centering
\includegraphics[width=\textwidth]{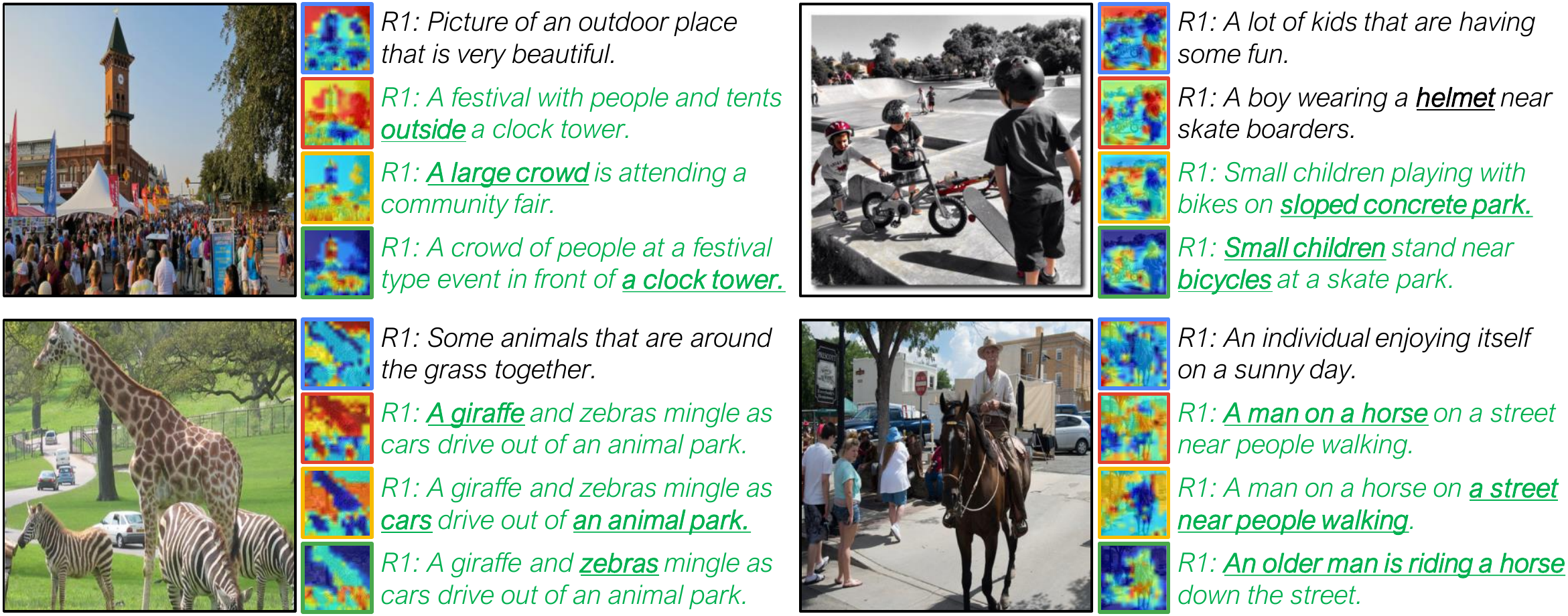}
\caption{For each element of the image embedding set, we present its attention map and the caption nearest to the element in the embedding space. Matching captions are colored in green. 
Entities corresponding to the attention maps are underlined.
}
\label{fig:qual_attn}
\end{figure*}

\noindent \textbf{Set prediction module:}
In every experiment, we set $T$ to 4. 
When using convolutional visual features, $D_h$ is set to 1024 and otherwise set to 2048. 

\noindent \textbf{Similarity and loss function:}
For smooth-Chamfer similarity, we set the scaling parameter $\alpha$ to 16.
In $\mathcal{L}_{\mathrm{tri}}$, we use margin $\delta$ of 0.1 for convolutional visual feature and 0.2 for ROI visual feature.
We multiply the factor of $0.01$ to $\mathcal{L}_{\mathrm{div}}$ and $\mathcal{L}_{\mathrm{mmd}}$.

\noindent \textbf{Training:}
The model is trained with the AdamW optimizer~\cite{adamw}. 
We construct the input batch with 200 images 
and all of their captions.
When using ROI visual features, the model is trained for 80 epochs with the initial learning rate of 1e-3, which is scheduled with cosine annealing~\cite{loshchilov2016sgdr}.
For the models employing convolutional visual features, we follow the training setting of previous work~\cite{song2019polysemous, Chen2021gpo}, which will be discussed in detail in the supplementary material.

\subsection{Comparisons with Other Methods}
Results on COCO and Flickr30K are summarized in Table~\ref{tab:coco_comparison} and Table~\ref{tab:f30k_comparison}, respectively.
For fair comparisons with previous work, our method is evaluated under three different visual extractors:
ResNet-152~\cite{resnet} pretrained on ImageNet~\cite{Imagenet},
ROI features~\cite{anderson2018bottom} pre-extracted by Faster R-CNN~\cite{faster_rcnn},
and ResNeXt-101~\cite{xie2017aggregated} pretrained on the Instagram~\cite{mahajan2018exploring} dataset.
For ResNet-152 and ResNeXt-101, input image resolution is set to respectively $224 \times 224$ and $512 \times 512$, following previous work~\cite{song2019polysemous, Chen2021gpo}.
For the textual feature extractor, we use either bi-GRU or BERT.
The ensemble results are obtained by averaging the similarity scores of two models trained with different random seeds.

Our method outperforms previous methods in terms of RSUM in every setting except for Flickr30K with ROI visual features. 
Even in this setting, ours achieves the second-best.
Note that NAAF, currently the best in this setting, requires two orders of magnitude heavier computations than ours at inference since it relies on cross-attention, as will be shown in the next section.
The ensemble of ours with ResNeXt-101 and BERT clearly outperforms all existing records, improving the previous best RSUM by $7.2\%$p and $2.0\%$p on COCO 5K and Flickr30K.
Comparisons with PVSE and PCME demonstrate the superiority of our set-based embedding framework. Our model outperforms both of them in every case. 
Specifically, our model improves RSUM by $3.3\%$p on COCO 5K and $10.0\%$p on Flickr30K, compared to PVSE.
Also, our method shows better results than models involving cross-attention networks~\cite{lee2018stacked,chen2020imram,Diao2021SGRAF}, 
which impose substantial computation as noted.
We finally emphasize that ours using a single model often outperforms previous work using ensemble~\cite{lee2018stacked,li2019visual,chen2020imram,Diao2021SGRAF}.

\subsection{Computation Cost Analysis}
\label{subsec:exp_complexity}

To demonstrate the efficiency of our model during evaluation time, we measure FLOPs required for computing a similarity score between representations of an image and a caption.
Ours demands $16.4$K FLOPs, while SCAN~\cite{lee2018stacked}, a representative cross-attention method, requires $1.24$M FLOPs.
This result shows that our method demands about $80$ times less floating-point operations than SCAN since the cross-attention requires re-processing of the image or caption representations for each query.
Also, it is worth noting that our method outperforms SCAN in terms of RSUM, as shown in Table \ref{tab:coco_comparison} and \ref{tab:f30k_comparison}.

\subsection{Analysis of Attention}
In Figure~\ref{fig:qual_attn}, attentions from the last aggregation block in the $f^\mathcal{V}$ are visualized. 
Each attention is used to encode individual elements of the image embedding set.
For each element, its nearest caption in the embedding space is presented together.
Thanks to slot attention, the set prediction module produces heterogeneous attention maps that focus on individual objects or contexts, which enable retrieving diverse captions.
In particular, on the top-right, elements are matched with the captions that describe different semantics (helmet, children, park, and bicycle).
Interestingly, we observed that one of the element slots attends to regions left out after other slots capture distinctive semantics, which wholly charges for retrieving related but highly abstract captions.

\subsection{Ablation Study}
We perform ablation studies to investigate the contributions of the proposed similarity function and set prediction module, in terms of RSUM and within-set variance.
For all ablation experiments, we use the models employing ROI visual features on Flickr30K.

\noindent \textbf{Importance of set similarity function:}
In the left of Table~\ref{tab:ablation}, we ablate the proposed similarity function and replace it with the MP, MIL, and Chamfer similarity. 
The results show that the models trained with smooth-Chamfer similarity surpass those trained with MIL or MP similarity under the same set prediction architecture.
Though Chamfer similarity improves the performance by resolving set collapsing, 
smooth-Chamfer similarity further improves performance by providing dense supervision during training.

\noindent \textbf{Importance of set prediction module:}
We also ablate the proposed set prediction module and substitute it with the PIE-Net, which is the baseline set prediction architecture of the previous set-based embedding methods, PVSE and PCME.
The results in the left of Table~\ref{tab:ablation} suggest that the impact of our set prediction module is significant.

\noindent \textbf{Verification of set collapsing:} 
In the right of Table~\ref{tab:ablation}, we report the average circular variance of embedding set.
Formally, the circular variance of embedding set $\mathbf{S}$ is denoted as $1 - \Vert\sum_{x \in \mathbf{S}}x/|\mathbf{S}|\Vert_{2}$.
As discussed earlier, results show that employing PIE-Net or MP leads to set collapsing.
Moreover, we report the result when aggregation block is replaced to transformer.
This variant results in a 13$\%$ lower variance on a linear scale and inferior performance since it does not guarantee disentanglement between elements.

\begin{table}[t!]
    \begin{subtable}[h]{0.48\linewidth}
        \centering
        \scalebox{0.64}{
        \begin{tabular}{lcc}
        \toprule
        Similarity      & Arch.         & RSUM                      \\ \midrule
        MIL             & Ours          & 491.7                     \\
        MP              & Ours          & 490.5                     \\
        Ours (Chamfer)        & Ours          & 499.6                     \\
        Ours (S-Chamfer)       & PIE-Net       & 483.3                     \\
        \ccol Ours (S-Chamfer) & \ccol Ours    & \ccol\textbf{500.8}       \\ 
        \bottomrule
        \end{tabular}
        }
     \end{subtable}
    \hfill
    \begin{subtable}[h]{0.51\linewidth}
        \centering
        \scalebox{0.76}{
        \begin{tabular}{lcc}
        \toprule
        Setting                    & $\log$(Var.)    & RSUM               \\ \midrule
        PIE-Net                     & -7.35           & 483.3              \\
        Ours \textbackslash w MP    & -5.27           & 490.5              \\
        Transformer                 & -2.27           & 496.1              \\
        \ccol Ours                  & \ccol -2.13     & \ccol \textbf{500.8}        \\
        \bottomrule
        \end{tabular}
       }
    \end{subtable}
    \caption{
        Ablation studies of the proposed similarity function and set prediction module.
    }
    \vspace{-3mm}
    \label{tab:ablation}
\end{table}
\begin{table}[t!]
    \footnotesize{
    \begin{subtable}[h]{\linewidth}
            \centering
            \scalebox{0.97}{
            \begin{tabular}{l|ccccccc}
            \toprule
            $K$     & 1       & 2        & 3     & 4               & 5     & 6     \\ \midrule
            RSUM    & 492.6   & 495.5    & 497.4 & \textbf{500.8}  & 498.4 & 499.3     \\
            \bottomrule
            \end{tabular}}
        \end{subtable}}
    \vspace{.5mm}

    \footnotesize{
    \begin{subtable}[h]{\linewidth}
        \centering
        \scalebox{0.97}{
        \begin{tabular}{l|cccccc}
        \toprule
        $T$     & 1       & 2        & 3     & 4               & 5     & 6       \\ \midrule
        RSUM    & 492.6   & 497.8    & 499.3 & \textbf{500.8}  & 499.4 & 499.1     \\
        \bottomrule
        \end{tabular}}
    \end{subtable}}
    \vspace{.5mm}
    
    \footnotesize{
    \begin{subtable}[h]{\linewidth}
        \centering
        \scalebox{0.85}{
        \begin{tabular}{l|ccccccc}
        \toprule
        $\alpha$    & 1       & 2        & 4     & 8      & 16     & 32    & 64    \\ \midrule
        RSUM        & 495.2   & 497.2    & 497.7 & 498.9  & \textbf{500.8}  & 499.3 & 499.0      \\
        \bottomrule
        \end{tabular}}
    \end{subtable}}
    \caption{Impact of hyperparameters for the proposed similarity function and set prediction module.
    }
    \vspace{-5mm}
     \label{tab:hyperparams}
\end{table}

\subsection{Impact of Hyperparameters}
In Table~\ref{tab:hyperparams}, we report the RSUM of the model while varying the cardinality of embedding sets $K$, the total number of iterations $T$, and the scaling parameter $\alpha$.
Results demonstrate that the model shows consistently high RSUM when $K > 3$, $T > 2$, and $\alpha > 4$.
Specifically, we notice that when $K=1$, which is identical to using a single embedding vector, the model shows significantly lower accuracy compared to others.
It suggests that embedding sets enable a more accurate retrieval by addressing the semantic ambiguity.
When $T = 1$, we observe substantial degradation of accuracy, which underpins that progressive refinement done by multiple aggregation blocks helps element slots aggregate semantic entities, as shown in Figure~\ref{fig:overview}.

\subsection{Analysis of Embedding Set Elements}
For further analysis of the embedding set elements, we use only one of the elements during the evaluation of the model.
Let $\mathbf{S}(i)$ be an element of $\mathbf{S}$ produced by the $i$-th element slot.
Table~\ref{tab:elements} summarizes the accuracy of the model in terms of RSUM when elements of $\mathbf{S}^\mathcal{V}$ and $\mathbf{S}^\mathcal{T}$ are ablated during evaluation.
Results present that using only one of the elements degrades accuracy.
Specifically, using only $\mathbf{S}^\mathcal{V}(2)$ during evaluation leads to far lower accuracy compared to using the other elements.
To check whether $\mathbf{S}^\mathcal{V}(2)$ is just a noisy element that hinders final performance, we also report RSUM when only $\mathbf{S}^\mathcal{V}(2)$ is ablated.
However, interestingly, we still observed the degraded accuracy compared to using a complete embedding set.
Though one of the elements often retrieves non-matching samples, the results demonstrate that using them together during evaluation helps the model find accurate matching samples.

We hypothesize that $\mathbf{S}^\mathcal{V}(2)$ is trained to capture highly ambiguous semantics.
Therefore, using them alone during evaluation leads to less accurate retrievals, while it improves performance when used together by successfully representing ambiguous situations and context.
A similar tendency could be observed in Figure~\ref{fig:qual_attn}, where one of the slots often encodes an element that is located nearby to a related but highly ambiguous caption in the embedding space.

\begin{table}[t!]
\centering
\begin{subtable}[h]{0.465\linewidth}
        \scalebox{0.65}{
        \setlength{\tabcolsep}{0pt}
        \begin{tabular}{@{}C{1.1cm}@{}@{}C{1.1cm}@{}@{}C{1.1cm}@{}@{}C{1.1cm}@{}|@{}C{1.2cm}@{}}
        \toprule
        \multicolumn{4}{c|}{Evaluation} & \\
        
        $\mathbf{S}^\mathcal{V}(1)$ & $\mathbf{S}^\mathcal{V}(2)$ &$\mathbf{S}^\mathcal{V}(3)$ & $\mathbf{S}^\mathcal{V}(4)$ & RSUM \\ \midrule
        
        \ccol\bcmark& \ccol\bcmark &\ccol\bcmark& \ccol\bcmark & \ccol\textbf{500.8} \\
        \bcmark & & & & {491.1} \\
        & \bcmark & & & {309.6} \\
        & & \bcmark & & {484.9} \\
        & & &\bcmark & {486.0} \\ \midrule
        \bcmark & & \bcmark &\bcmark & {500.2} \\ \bottomrule
        \end{tabular}
        }
\end{subtable}
\hfill
\begin{subtable}[h]{0.525\linewidth}
        \scalebox{0.75}{
        \setlength{\tabcolsep}{0pt}
        \begin{tabular}{@{}C{1.1cm}@{}@{}C{1.1cm}@{}@{}C{1.1cm}@{}@{}C{1.1cm}@{}|@{}C{1.2cm}@{}}
        \toprule
        \multicolumn{4}{c|}{Evaluation} & \\
        $\mathbf{S}^\mathcal{T}(1)$ & $\mathbf{S}^\mathcal{T}(2)$ &$\mathbf{S}^\mathcal{T}(3)$ & $\mathbf{S}^\mathcal{T}(4)$ & RSUM \\ \midrule
        
        \ccol{\bcmark}& \ccol\bcmark &\ccol\bcmark& \ccol\bcmark & \ccol\textbf{500.8} \\
        \bcmark & & & & {481.9} \\
        & \bcmark & & & {483.0} \\
        & & \bcmark & & {481.7} \\
        & & &\bcmark & {497.2} \\ \bottomrule
        
        \end{tabular}
        }
\end{subtable}
\caption{
RSUM on the Flickr30K dataset when elements of the $\mathbf{S}^\mathcal{V}$(\textit{left}) and $\mathbf{S}^\mathcal{T}$(\textit{right}) are ablated during evaluation.
}
\vspace{-3mm}
\label{tab:elements}
\end{table}

\section{Conclusion}
\label{sec:conclusion}
We propose the novel set embedding framework for cross-modal retrieval, consisting of the set prediction module and smooth-Chamfer similarity.
The proposed set prediction module outputs an embedding set that successfully captures ambiguity, while smooth-Chamfer similarity resolves undesirable effects of the existing similarity functions.
As a result, our model surpasses most of the previous methods, including methods involving higher computation costs, on COCO and Flickr30K datasets.
We will extend our model for the different modalities and tasks in the future.

\small{
\noindent \textbf{Acknowledgement.}%
This work was supported by the NRF grant and %
the IITP grant %
funded by Ministry of Science and ICT, Korea 
(NRF-2018R1A5-A1060031--20\%, %
NRF-2021R1A2C3012728--50\%, %
IITP-2019-0-01906--10\%, %
IITP-2022-0-00290--20\%).%
}

\clearpage
\small{
\bibliographystyle{ieee_fullname}
\bibliography{cvlab_kwak}
}

\appendix

\twocolumn[
  \begin{@twocolumnfalse}
    \huge{\textbf{Appendix}}
    \vspace{5mm}
  \end{@twocolumnfalse}
]

\addcontentsline{toc}{section}{Appendices}
\renewcommand\thefigure{A\arabic{figure}}
\renewcommand{\thetable}{A\arabic{table}}
\setcounter{figure}{0}
\setcounter{table}{0}

\normalsize{
This supplementary material provides details and additional results of our method that have not been presented in the main paper due to the page limit.
In Section~\ref{sec:supp_detail}, we first provides further implementation details for each feature extractor settings.
We present additional experimental results including in-depth analysis of the model, results on the additional benchmarks, more ablation studies, and embedding space visualization in Section~\ref{sec:supp_exp}.
Finally, Section~\ref{sec:supp_quals} presents more qualitative results for the set prediction module.

\section{Implementation Details}
\label{sec:supp_detail}
Our model is implemented with PyTorch~\cite{pytorch} v 1.8.1. 
Automatic mixed precision is used for faster and more efficient training.
In addition to implementation details provided in the main paper, training settings vary based on feature extractors.

\noindent \textbf{ResNet-152 + bi-GRU:}
In this setting, the model is trained for 120 epochs with the initial learning rate of 1e-3 and 2e-3 on COCO and Flickr30K, respectively.
The learning rate for the set prediction module is scaled by $0.1$ and $0.01$ on COCO and Flickr30K, respectively.
The learning rate decays by a multiplicative factor of $0.1$ for every 10 epochs.
Following~\cite{song2019polysemous}, the CNN is not trained for the first 50 epochs.
Training is performed using a single RTX 3090 GPU.

\noindent \textbf{Faster-RCNN + bi-GRU:}
The learning rate for the set prediction module is scaled by $0.1$ and $0.05$ on COCO and Flickr30K, respectively. 
Following~\cite{Chen2021gpo}, we drop 20\% of ROI features and words during training.
Training is performed using a single RTX 3090 GPU.

\noindent \textbf{ResNeXt-101 + BERT:}
In this setting, we construct a batch with 128 images and their entire matching captions. 
The model is trained for 50 epochs with the initial learning rate of 1e-4, where the learning rate decays by a multiplicative factor of $0.1$ for every 20 epochs.
Following~\cite{Chen2021gpo}, the learning rate for CNN is scaled by $0.1$. 
The statistics of the batch normalization~\cite{Batchnorm} layer are fixed during training. 
The CNN is not trained in the first epoch. In the first epoch, triplet loss without mining is used, whereas the hardest negative mining is used for later epochs.
Training is performed using two A100 PCIe GPU.

\section{Additional Experiments}
\label{sec:supp_exp}

\subsection{In-Depth Computation Cost Analysis}
\label{subsec:supp_cost}
In addition to the computation cost analysis presented in the main paper, we compare our method with SCAN~\cite{lee2018stacked} and VSE$_\infty$~\cite{Chen2021gpo}, focusing on FLOP and latencies.
FLOPs and latencies are measured during computing similarity score between data and then obtaining nearest top-10 retrieval result on Flickr30K validation.
Given the cost of VSE$_\infty$ as 1 in terms of FLOPs, our method has an approximate cost of 16, while SCAN has a cost of 1,280.
VSE$_\infty$, Ours, and SCAN have latencies of 159ms, 168ms, and 198,121ms, respectively.
While our model is substantially more efficient than cross-attention based methods like SCAN, it demands more computation than single embedding methods such as VSE$_\infty$.

Nevertheless, we observed that when increasing the embedding dimension of VSE$_\infty$ to match the FLOPs of ours, it results in performance drop of 10.8\%p on Flickr30K RSUM.
This finding indicates that the improvement we achieved is not merely due to the additional computation, but rather stems from our set-based embedding approach.

\subsection{Results on ECCV Caption and CrissCrossed Caption}
\label{subsec:supp_eccv}
\begin{table}[t!]
\centering
\setlength{\tabcolsep}{3.5pt}
\scalebox{0.8}{

\begin{tabular}{l|cccc|cccc}
\toprule
& \multicolumn{4}{c|}{Image-to-text} & \multicolumn{4}{c}{Text-to-image} \\ 
& \multicolumn{3}{c}{ECCV Caption} & CxC & \multicolumn{3}{c}{ECCV Caption} & CxC \\
& mAP@R & R-P & R@1 & R@1 & mAP@R & R-P & R@1 & R@1 \\ \midrule
VSRN& 30.8 & 42.9 & 73.8 & {55.1} & \textbf{53.8} & \textbf{60.8} & {89.2} & {42.6} \\
VSE$_\infty$ & \underline{34.8} & \underline{{45.4}} & \underline{{81.1}} & \underline{{67.9}} & {50.0} & {57.5} & \textbf{91.8} & \underline{53.7}\\ \midrule
Ours & \textbf{36.0} & \textbf{46.4} & \textbf{84.7} & \textbf{72.3} & \underline{{51.0}} & \underline{58.5} & \underline{91.6} & \textbf{55.5}\\
\bottomrule
\end{tabular}
}
\caption{\small mAP@R, R-Precision, and Recall@1 are reported for both ECCV Caption and CrissCrossed Caption (CxC). Results on image-to-text retrieval and text-to-image retrieval are reported.
}
\vspace{-5mm}
\label{tab:supp_eccv}
\end{table}
Recently, benchmarks for the cross-modal retrieval, such as CrissCrossed Caption (CxC)~\cite{Parekh2020CrisscrossedCE} and ECCV Caption~\cite{Chun2022ECCVCC}, have been proposed to address the missing correspondences issue in conventional benchmarks.
In particular, within the COCO dataset, each caption associated with only one image, while each image is matched with five different captions.
This missing correspondence leads to a large numbers of false-negatives, as captions that may accurately describe other images are overlooked during testing, thus obstructing the correct evaluation of the models.
CxC and ECCV caption mitigate the false-negative issue by introducing re-established correspondences between images and captions in the COCO test split.

For the comprehensive evaluation of our model, we report the results of our best model on CxC and ECCV caption, in the Table~\ref{tab:supp_eccv}.
We compare our method with VSRN~\cite{li2019visual} and VSE$_\infty$~\cite{Chen2021gpo}, which are reported to achieve previous best results on ECCV caption and CxC, respectively~\cite{Chun2022ECCVCC}.
It is important to note, however, that VSRN is one of the machine annotators used to construct the ECCV Caption dataset itself, which may introduce potential machine bias into the dataset and result in inflated evaluations.
Despite this, our work achieves the best or second-best performance in every metric, which is particularly noteworthy given that the results on ECCV caption are known to have a low correlation with those on conventional benchmarks.

\subsection{Ablation Study of Architectural Modification to Slot Attention}
\label{subsec:supp_slot}
As described in the Section~2 of the main paper, we made three modifications to the original slot attention~\cite{locatello2020object}: (1) using learnable embeddings for initial element slots, (2) replacing GRU~\cite{cho2014learning} with a residual sum, and (3) adding a global feature into the final element slots. 
Without these modifications, training failed, yielding a COCO 5K RSUM of 0.74. 
Ablations of (1), (2), and (3) result in COCO 5K RSUM of 427.2 (-3.5\%p), 423.3 (-7.4\%p), and 342.0 (-88.7\%p), respectively.
It is evident that (3) has the most substantial impact on retrieval performance. 
This is because a global feature effectively addresses samples with little ambiguity, particularly during the early stages of training when addressing semantic ambiguity is challenging for the network. 
However, this does not imply that global features dominate the embedding set, as verified by the high circular variance in Table~3 of the main paper.

\subsection{$t$-SNE Visualization of Embedding Space}
\label{subsec:supp_tsne}

\begin{figure}[t!]
    \centering
    \includegraphics[width=0.98\linewidth]{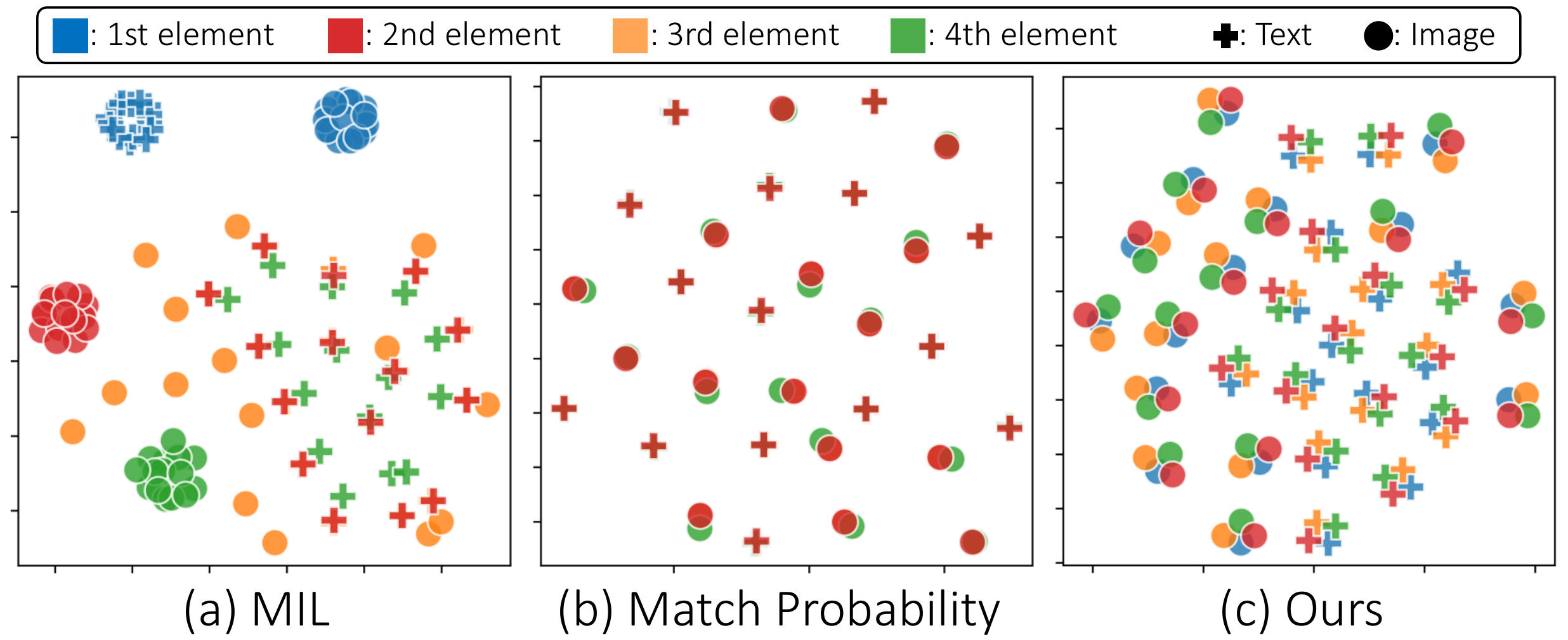}
\caption{
$t$-SNE visualization of the embedding spaces.
Visualized marker represent elements of embedding sets.
Color and shape of the markers denote corresponding slots and modality of the data, respectively.
}
\vspace{-5mm}
\label{fig:supp_tsne}
\end{figure}
Figure~\ref{fig:supp_tsne} visualizes embedding spaces of our model trained with different similairty functions through $t$-SNE.
Each marker represents an embedding set element, and its color and shape indicate the corresponding slot and modality, respectively.
The visualization shows two side effects of MIL and MP: sparse supervision and set collapsing.
MIL leaves some slots untrained, which is observed as clusters of elements produced from the same slots.
Conversely, MP suffers from the set collapsing, making it difficult to distinguish set elements in $t$-SNE, as also verified with the small within-set circular variance in Table 3 of the main paper.
Unlike MIL and MP, our smooth-Chamfer similarity enables learning of a model that takes account of every set element, maintaining sufficient within-set variance to encode semantic ambiguity.

\section{Additional Qualitative Results}
\label{sec:supp_quals}

In Figure~\ref{fig:supp_attn1} and Figure~\ref{fig:supp_attn2}, we present additional visualization of attention map from the visual set prediction module $f^\mathcal{V}$. Visualizations of attention maps, including ones presented in the main paper, are obtained from the model using ResNeXt + BERT feature extractors.
Attention maps from each iteration are presented together, where $t=4$ is the last iteration.
For each attention map from the last iteration, its corresponding element of embedding set and nearest caption are provided together.
Results show that the aggregation block produces heterogeneous attention maps capturing various semantics, such as different objects (1st row of Figure~\ref{fig:supp_attn1}) and action (2nd row of Figure~\ref{fig:supp_attn2}).
Moreover, in every case, we can observe that element slots are progressively updated to capture distinctive semantics, starting from sparse and noisy attention maps. 

Specifically, in Figure~\ref{fig:supp_attn2}, we present the examples where the nearest captions of multiple elements are the same.
For instance, in the 2nd row of Figure~\ref{fig:supp_attn2}, each element attends to individual entities (sky, larger giraffe, grassy area, and baby giraffe), but their nearest captions, which describe the entire scene, are the same.
Results imply that by fusing element slots with the global feature, elements of the embedding set can preserve the global context while focusing on distinctive semantics.
These characteristics help model when samples with little ambiguity are given, such as a caption describing the entire scene (2nd row of Figure~\ref{fig:supp_attn2}) or an image containing a single iconic entity (3rd row of Figure~\ref{fig:supp_attn2}).

}

\clearpage
\begin{figure*} [!t]
\centering
\includegraphics[width=\textwidth]{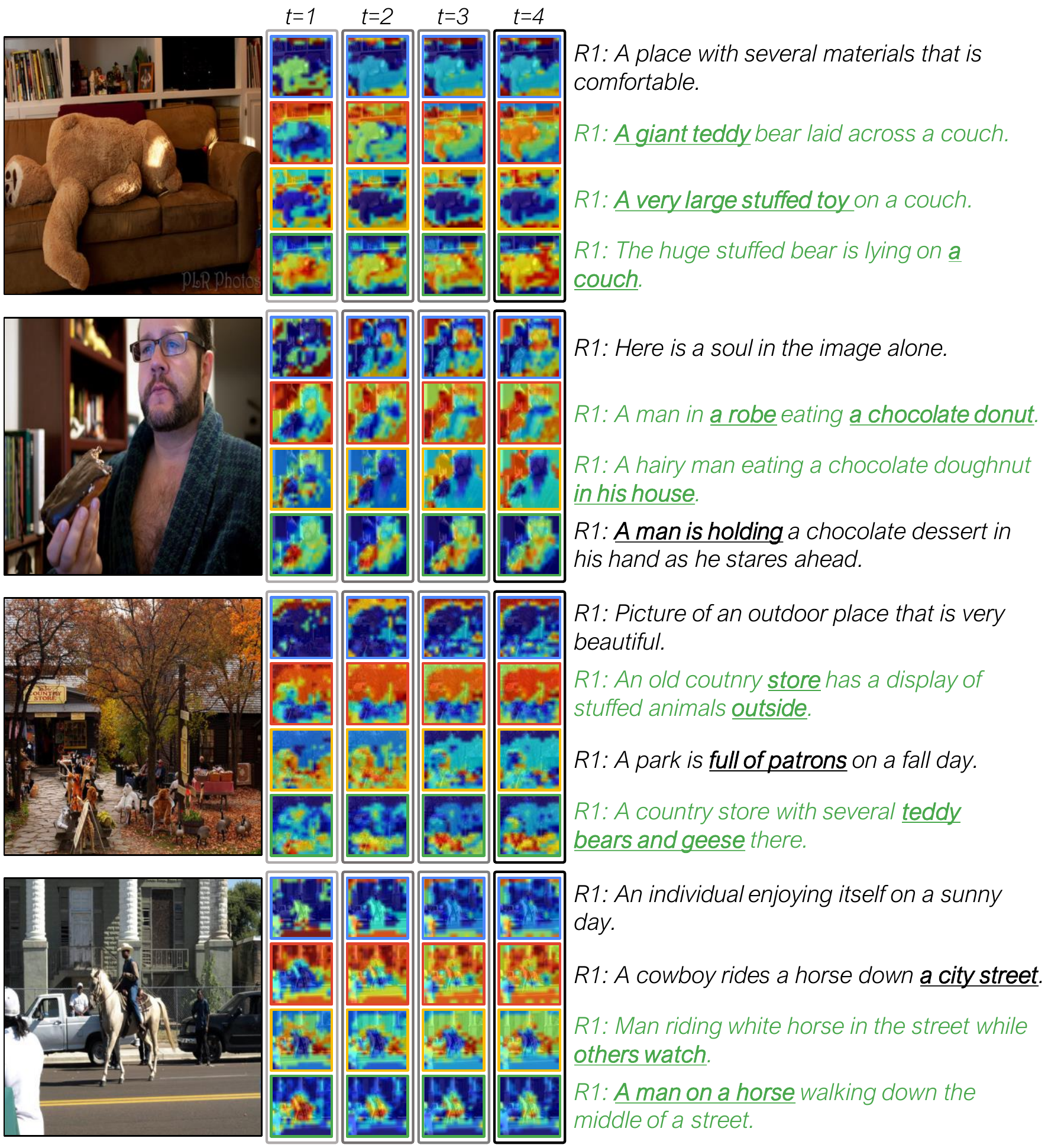}
\caption{{For each element of the image embedding set, we present its attention map and the caption nearest to the element in the embedding space. Matching captions are colored in green. 
Entities corresponding to the attention maps are underlined.}
} 
\label{fig:supp_attn1}
\end{figure*}
\clearpage
\begin{figure*} [!t]
\centering
\includegraphics[width=\textwidth]{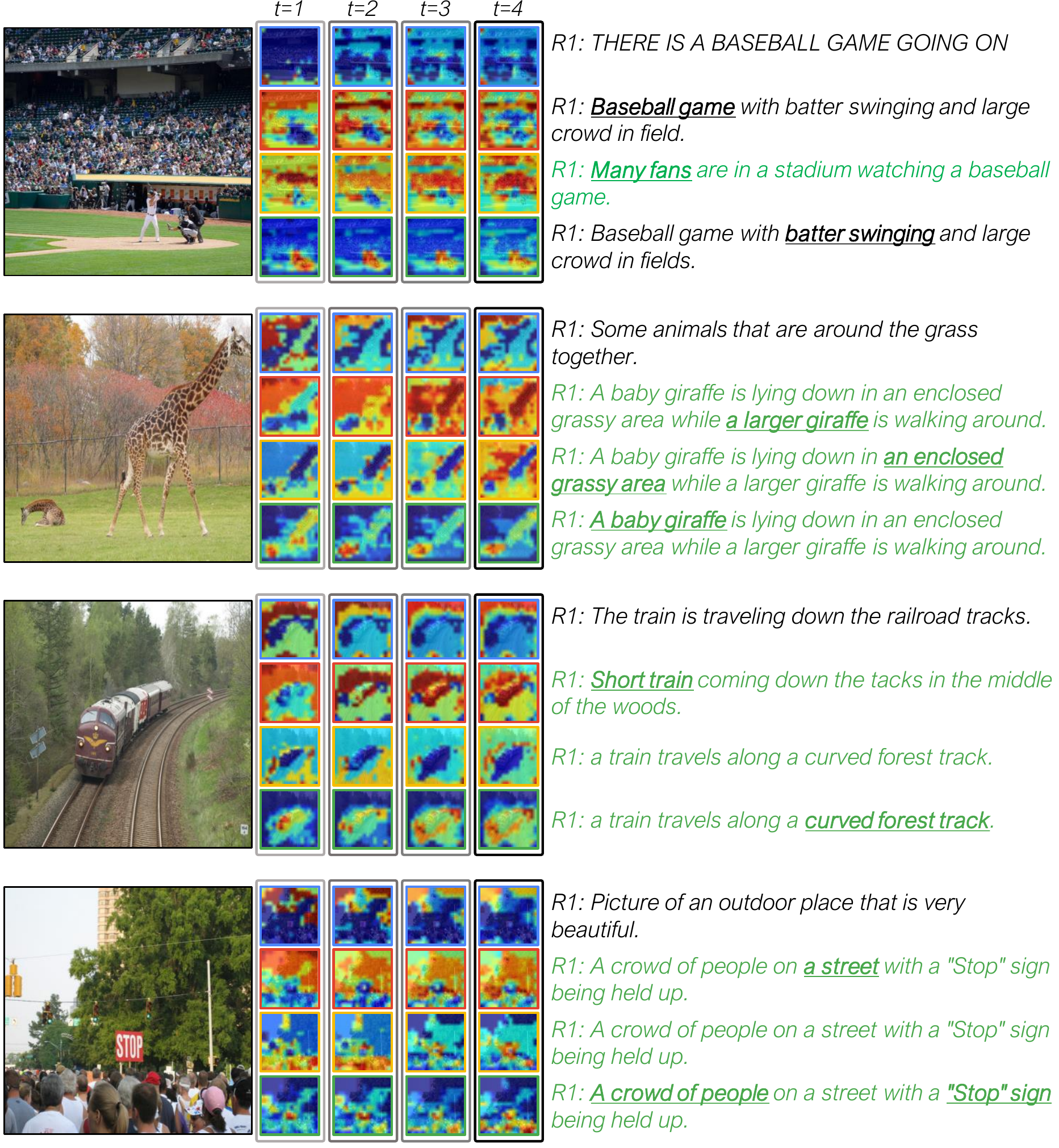}
\caption{{For each element of the image embedding set, we present its attention map and the caption nearest to the element in the embedding space. Matching captions are colored in green. 
Entities corresponding to the attention maps are underlined.}
} 
\label{fig:supp_attn2}
\end{figure*}

\pagebreak

\end{document}